\documentclass[runningheads]{llncs}
\usepackage[T1]{fontenc}
\usepackage{graphicx}
\usepackage{amsmath}
\usepackage{amsfonts}
\usepackage[ruled,vlined]{algorithm2e}
\usepackage{xcolor}
\usepackage{mdframed}
\usepackage{multirow}
\usepackage{multicol}
\usepackage{hyperref}
\usepackage{color}
\usepackage{tabularx}
\usepackage{orcidlink}
\usepackage{marvosym}
\usepackage{etoolbox}
\AtBeginEnvironment{thebibliography}{\scriptsize}

\begin{document}
%
% \title{Contribution Title\thanks{Supported by organization x.}}
\title{Unifying Visual and Semantic Feature Spaces with Diffusion Models for Enhanced Cross-Modal Alignment}
\author{Yuze Zheng~\orcidlink{0009-0003-5718-8366} \and
Zixuan Li~\orcidlink{0009-0005-2971-9478} \and
Xiangxian Li~\orcidlink{0000-0001-6638-2361} \and
Jinxing Liu~\orcidlink{0009-0009-4682-3268} \and
Yuqing Wang~\orcidlink{0000-0002-4151-8290} \and
Xiangxu Meng~\orcidlink{0000-0001-7290-5659} \and
Lei Meng\textsuperscript{\Letter}~\orcidlink{0000-0002-0273-5946}}
\institute{School of Software, Shandong University, Jinan, China}
% \url{https://ercdm.sdu.edu.cn/info/1013/1523.htm}}
%
\maketitle              % typeset the header of the contribution
\begin{abstract}

Image classification models often demonstrate unstable performance in real-world applications due to variations in image information, driven by differing visual perspectives of subject objects and lighting discrepancies. To mitigate these challenges, existing studies commonly incorporate additional modal information matching the visual data to regularize the model's learning process, enabling the extraction of high-quality visual features from complex image regions. Specifically, in the realm of multimodal learning, cross-modal alignment is recognized as an effective strategy, harmonizing different modal information by learning a domain-consistent latent feature space for visual and semantic features. However, this approach may face limitations due to the heterogeneity between multimodal information, such as differences in feature distribution and structure. To address this issue, we introduce a Multimodal Alignment and Reconstruction Network (MARNet), designed to enhance the model's resistance to visual noise. Importantly, MARNet includes a cross-modal diffusion reconstruction module for smoothly and stably blending information across different domains. Experiments conducted on two benchmark datasets, Vireo-Food172 and Ingredient-101, demonstrate that MARNet effectively improves the quality of image information extracted by the model. It is a plug-and-play framework that can be rapidly integrated into various image classification frameworks, boosting model performance.

\end{abstract}
\section{Introduction}

Visual classification is a critical task in the field of computer vision\cite{qizhuang-mm}\cite{wyq-ijcnn}\cite{zitan-mm}\cite{jingyu-icmr}. However, the quality of visual images is susceptible to various factors, including but not limited to, interference from non-main elements and changes in lighting angles, leading to inconsistent performance in image classification\cite{yuqing-eccv}\cite{yuqing-caai}\cite{xiangxian-cvm}. With the rapid development of social media platforms, a vast amount of textual information related to visual images has emerged. These pieces of information present a complex relationship of mutual dependence and complementarity, which can compensate for the shortcomings of single-modal information. However, the potential complementarity between images and texts is often limited due to the fundamental differences between these two forms of information, thus affecting the effectiveness of multimodal learning\cite{mm19}\cite{multi_learning}. Therefore, effectively integrating and utilizing cross-modal data information becomes key to enhancing the performance of multimodal learning.

In recent years, researchers in multimodal learning have commonly adopted cross-modal representation alignment strategies to reduce the heterogeneity between different modal information. These strategies can be broadly divided into two categories: those based on distance metrics \cite{ssan}\cite{swd}\cite{cdd}and those based on contrastive learning\cite{ita}\cite{sdm}\cite{tema}. Distance-based alignment methods mainly constrain the spatial distance between different sources of information, such as category center distance or decision space distance, to effectively mitigate the problem of distance differences between modal information in the information space. In contrast, contrastive learning-based alignment methods divide multimodal information into positive and negative samples and enhance the similarity between positive samples while separating them from negative samples by comparing sample similarities. This approach strengthens the distinction and interactivity of information in the representation space. However, both methods tend to focus on the distance between representations while aligning cross-modal representations as much as possible, neglecting the significant distribution differences between different modal representations, which is a challenge that needs to be addressed in multimodal learning.

\begin{figure}
\centering
 \includegraphics[width=\textwidth]{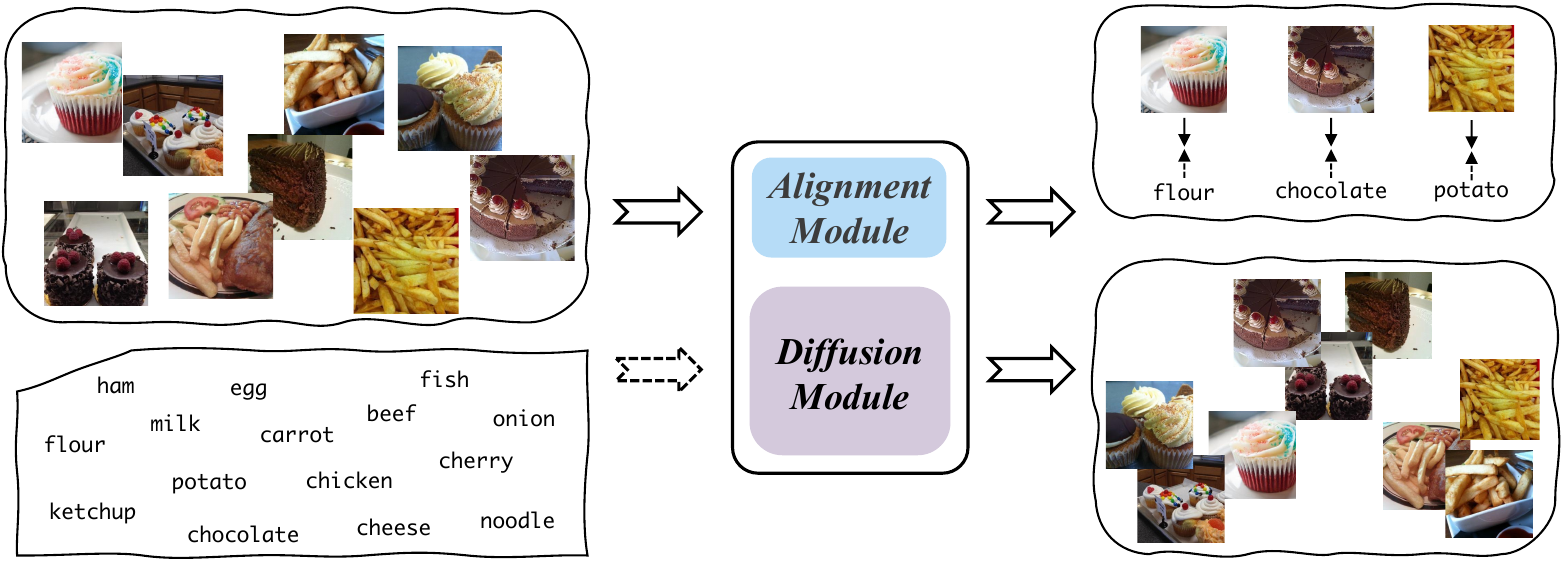}
\caption{The proposed MARNet schematic diagram. The network mainly comprises alignment and diffusion modules, wherein the alignment module matches and aligns image and text information, and the diffusion module reconstructs the distribution of image information.} \label{motivation}
\vspace{-0.2cm}
\end{figure}

To deeply address the challenges in cross-modal representation alignment, this study first thoroughly analyzes the common algorithmic frameworks within the two categories of alignment methods, assessing their strengths and limitations. Based on this analysis, we introduce an innovative \textbf{m}ultimodal \textbf{a}lignment and \textbf{r}econstruction \textbf{net}work, named MARNet. As illustrated in Figure \ref{motivation}, MARNet, through a contrastive learning-based alignment strategy, effectively resolves the issue of representation confusion in the visual space while enhancing the separation of samples within the same category space. Unlike previous alignment methods, we designed a cross-modal diffusion reconstruction module to complement the deficiencies of traditional alignment methods. By introducing a diffusion model with guided conditions, we achieved deep interaction between visual and textual information, significantly optimizing the distribution of visual information and enhancing the model's perception of the visual information's core areas. Through this representation fusion strategy, we enabled two independent modules to complement each other, thereby achieving the goal of enhancing visual representation and improving model robustness.

In practical image-text multimodal classification tasks, we conducted extensive experimental validation of MARNet, especially on two single-label datasets involving dishes and ingredients, Vireo-Food172 and NUS-WIDE. Compared to previous alignment frameworks, MARNet, as a model-agnostic algorithmic framework, significantly enhanced the quality of visual representation and improved the framework's performance in downstream tasks through its unique embedding matching alignment module and cross-modal diffusion reconstruction module. Through case analysis, we further demonstrated how the cross-modal diffusion reconstruction module, by incorporating textual information, significantly improved the distribution of visual information in the representation space and strengthened the model's ability to recognize visual subjects, showcasing MARNet's strong potential and practical value in the field of multimodal learning. The innovative contributions of this chapter are mainly reflected in:
\begin{enumerate}
    \item We propose an innovative multimodal alignment and reconstruction network, named MARNet. Under the fine guidance of textual information, this network significantly improves the quality of visual information and markedly enhanced the model's decision-making ability in the visual domain. MARNet is designed with flexibility, allowing easy integration with current mainstream two-channel models to strengthen the capability of feature representation, demonstrating outstanding versatility and reusability.
    \item We introduce a cross-modal diffusion reconstruction module. This module utilizes a diffusion model to smoothly unfold multimodal data along the time axis, correcting visual modal information through deep interaction, effectively optimizing the aggregation distribution of similar visual information. This module not only enhances the model's ability to process visual information but also deepens its understanding of multimodal data.
    \item We explore the respective advantages and limitations of existing cross-modal alignment methods and diffusion model reconstruction representations, providing referential conclusions for future research.
\end{enumerate}

\section{Related Work}
\subsection{Cross-modal Alignment}
In common cross-modal learning scenarios, there is a clear distribution difference in the representation space among different modal data, and representations of the same category from different modalities are disorganized. Cross-modal representation alignment is needed to mitigate differences between cross-modal representations. Leading-edge cross-modal alignment methods can be divided into two paradigms: distance metric-based and contrastive learning-based methods.

In distance metric-based alignment methods, Lee \textit{et al.}\cite{swd} project the decision information from different modalities into a spherical space, and optimize the distance using the Wasserstein metric. Li \textit{et al.}\cite{ssan} propose centroid alignment, which explicitly pulls the distance of modal corresponding class closer by calculating the centroids of clusters, while also incorporating decision information for implicit alignment. Kang \textit{et al.}\cite{cdd} adopt a clustering alignment method, optimizing the distance within and between category clusters by constructing an intra-cluster sample matrix, achieving class-aware alignment.

In contrastive learning-based alignment methods, Jiang \textit{et al.}\cite{sdm} calculate the cosine similarity of visual and textual representations among positive samples, enhancing the similarity between modalities. Xie \textit{et al.}\cite{tema} integrate attention mechanisms into alignment methods on top of traditional global representation alignment, aligning internal information of representations at a finer granularity. Wang \textit{et al.}\cite{ita} improving upon the InfoNCE\cite{infoNCE} through contrastive learning, maximize the similarity of positive image-text pairs in a common representation space while minimizing the negative impact of other sample pairs.

\subsection{Diffusion Models for Representation Learning}

The diffusion model is inspired by non-equilibrium thermodynamics\cite{ddpm_org}. Ho \textit{et al.}\cite{ddpm} treat the diffusion process as a Markov chain by progressively adding random noise to the data. They train neural networks to learn the diffusion process, enabling them to denoise images corrupted with Gaussian noise. 

Currently, diffusion models are mostly applied to generative tasks\cite{ddpm_generate}\cite{lixSiggraphAsia}. In cross-modal diffusion models, there are commonly two approaches. One is using classifier-free guidance\cite{ddpm_classifier}, where text is used as a condition to guide image generation with noise. The other is simultaneously adding noise from multiple modalities into the network for multi-modal generation\cite{ddpm_generate_classifier}.

In terms of network structures used in diffusion models, U-Net architecture is commonly employed in the image domain for noise prediction, with intra-layer changes in image channels\cite{ddpm}. Additionally, some studies have utilized MLP structures for diffusion in user-item interactions without channel dimensions\cite{ddpm_rec}\cite{DDPM_IJCAI}, focusing on simpler feature transformations.

Regarding image classification tasks based on diffusion models, Li \textit{et al.}\cite{ddpm_zero} introduced a method to evaluate diffusion models as zero-shot classifiers. Clark \textit{et al.}\cite{ddpm_t2i} used density estimation calculated by a large-scale text-to-image generation model for zero-shot classification.

\section{Method}
\begin{figure}
\vspace{-0.2cm}
\includegraphics[width=\textwidth]{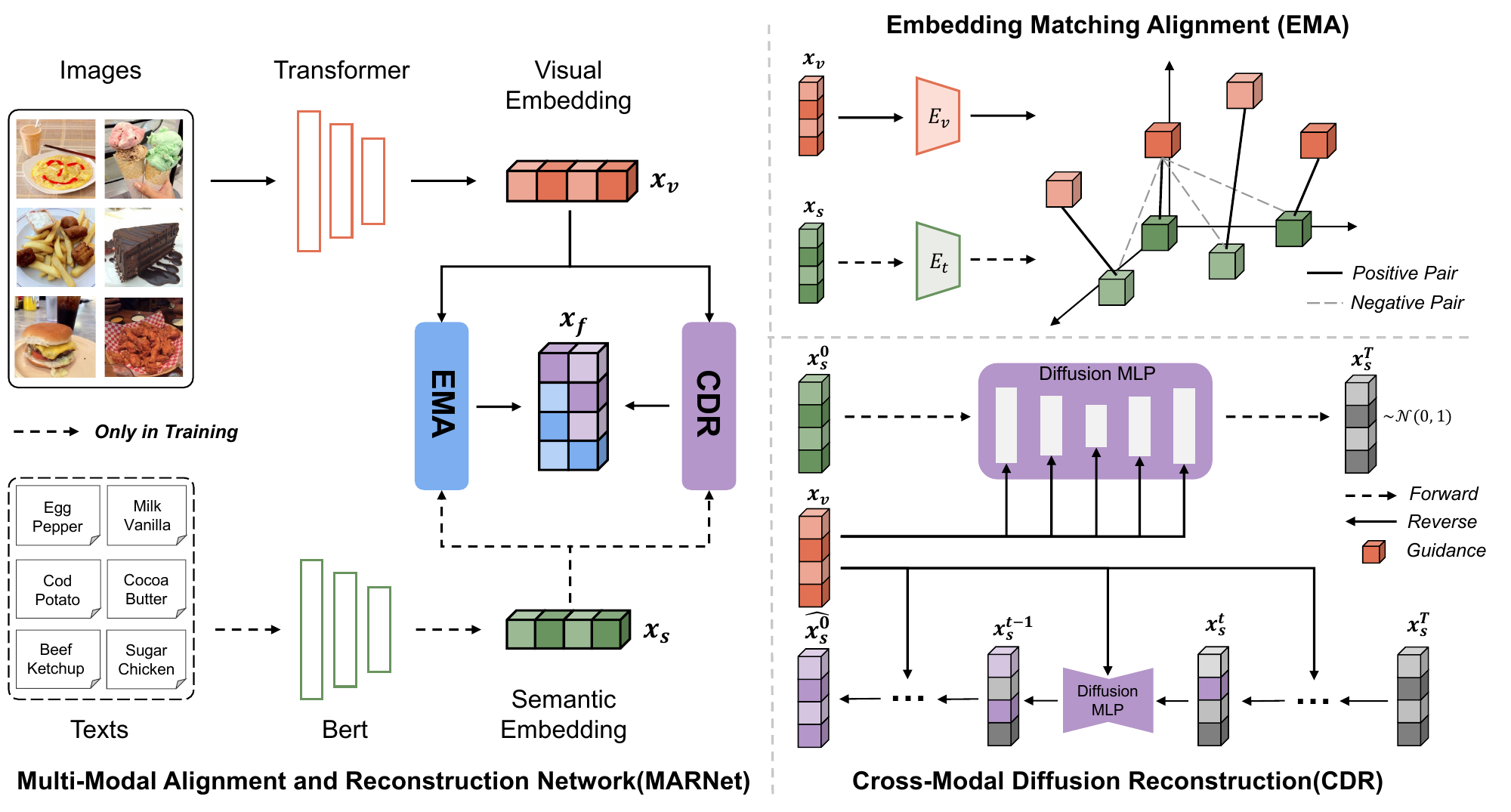}
\caption{The framework diagram of MARNet. The input to MARNet is image-text data pairs, which are processed by neural networks for vision and text to obtain $x_v$ and $x_s$, respectively. Modules EMA and CDR handle the multi-modal representations and output representations $x_{EMA}$ and $x_{CDR}$, which are fused in the end. } \label{framework}

\vspace{-1cm}
\end{figure}

% MARNet = Multi-Modal Representation Alignment and Reconstruction Network
In this section, we elaborate on how our proposed framework(MARNet), aligns image-text sample pairs in the representation space through \textbf{embedding matching alignment module(EMA)}, and how it mitigates the distribution differences existing in cross-modal information via \textbf{cross-modal diffusion reconstruction module(CDR)}, ultimately enhancing the interaction between cross-modal data.The overall architecture diagram of MARNet is shown in Figure ~\ref{framework}.

\subsection{Overall Approach}
Our goal is to learn more multi-dimensional and rich visual representations from paired image-text data through privileged information learning, in order to improve the classification performance of Multi-Modal Alignment and Reconstruction Network(MARNet). More specifically, we treat the precious and scarce ingredient data as privileged information to guide the representation of image data, that is, the training samples consist of $\mathcal{N}$ pairs of image-text data $\mathcal{S}_{N} = \{(p_1, i_1), (p_2, i_2), ... ,(p_n, i_n) \}$, while the test samples only contain $\mathcal{M}$ pieces of photo data $\mathcal{S}_{M} = \{p_1, p_2, ... ,p_m\}$. We use a visual encoder $\mathcal{F}_{v}$ to extract the representations of image data $\mathcal{R}_{v}=\{x_v^1, x_v^2, ..., x_v^n\}$,where $x_v = \mathcal{F}_{v}(p)$, and similarly for text data, $\mathcal{R}_{s}=\{x_s^1, x_s^2, ..., x_s^n\}$,where $x_s = \mathcal{F}_{s}(i)$. We take the visual representations $x_v$ and semantic representations $x_s$ as inputs for the subsequent two modules. 
In the embedding matching alignment module, we finely align the cross-modal representations through contrastive matching learning and generate representations $x_{EMA}$. In the cross-modal diffusion reconstruction module, we adopt an improved diffusion model to stably and smoothly infiltrate the visual representations $x_v$ into the semantic representations $x_s$ and sample to generate representations $x_{CDR}$ from Gaussian noise ${N}_{G}$. Finally, we fuse the output representations of the modules as $x_f$ and predict the final classification results.
\begin{align}
    x_{EMA} &= EMA(\mathcal{F}_{v}(p), \mathcal{F}_{s}(i))
    \\
    x_{CDR} &= CDR(\mathcal{F}_{v}(p), \mathcal{F}_{s}(i))
    \\
    x_f &= {fusion}(x_{EMA}, x_{CDR})
    \\
    \hat{C} &= {Classifier}(x_f)
\end{align}

% \vspace{-0.5cm}
\subsection{Embedding Matching Alignment}
% \vspace{-0.3cm}
% an instance-wise Image-Text
Based on the positive and negative sample matching alignment method of contrastive learning, we adopt an instance-wise  Alignment(ITA) approach\cite{ita}. This alignment method is an improvement based on InfoNCE\cite{infoNCE}, which calculates the matching similarity($Sim(x_v^i, x_s^i)$) of image-text representations in feature space within a batch as a constraint to align cross-domain information. When enhancing the similarity of a set of image-text representation pairs using positive and negative sample matching methods, it also reduces the matching degree of the visual representation $x_v^i$ with other semantic representations $x_s^j$, where $i \ne j$. Definition of cosine similarity is as follows:
\begin{equation}
    Sim(x_v^i, x_s^i) = \frac{{x_v^i} \cdot {x_s^i}} {||{x_v^i}|| ||{x_s^i}||}
\label{eq:cosine_sim}
\end{equation}
 where ${x_v^i} \cdot {x_s^i}$ is the dot product of vectors, and $||x_v^i|| ||x_s^i||$ is the product of the modulus of vectors.

We design two encoders $E_v$ and $E_s$, each consisting of a linear layer, and use an activation function $g(x)$ (\textit{i.e.}, LeakyReLU). The encoder $E_v$ is used to map visual representations $x_v$, while the encoder $E_s$ is used to map semantic representations $x_s$, both to the same feature space $\mathbb{R}^d$. 
\begin{align}
    g(x) &= \begin{cases}
    x, & \text{if } x \geq 0 \\
    \alpha x, & \text{otherwise}
\end{cases}
    \\
    x_{v^{'}} &= g(E_v(x_v)), x_{v^{'}} = \begin{bmatrix} x_{v^{'}}^1 \ x_{v^{'}}^2 \ \vdots \ x_{v^{'}}^d \end{bmatrix}
    \\
    x_{s^{'}} &= g(E_s(x_s)), x_{s^{'}}= \begin{bmatrix} x_{s^{'}}^1 \ x_{s^{'}}^2 \ \vdots \ x_{s^{'}}^d \end{bmatrix}
\end{align}
where $\alpha$ is set to the default value of 0.01. 

We use the mutual matching cosine similarity between visual representation $x_v^{'}$ and semantic representation $x_s^{'}$ as the alignment constraint $\mathcal{L}_{ITA}$.
\begin{align}
\mathcal{L}_{v2s} &= -\log \frac{\exp \left({Sim}\left(x_{v^{'}}^i, x_{s^{'}}^i\right) / \tau \right)}{\sum_{b=1}^{B} \exp \left({Sim}\left(x_{v^{'}}^i, x_{s^{'}}^b\right) / \tau \right)}
\\
\mathcal{L}_{s2v} &= -\log \frac{\exp \left({Sim}\left(x_{s^{'}}^i, x_{v^{'}}^i\right) / \tau \right)}{\sum_{b=1}^{B} \exp \left({Sim}\left(x_{s^{'}}^i, x_{v^{'}}^b\right) / \tau \right)}
\\
\mathcal{L}_{ITA} &= \mathcal{L}_{v2s} + \mathcal{L}_{s2v}
\end{align}
where $B$ is batch size, $\tau$ is a temperature factor, which is initialized as 0.07. The definition of cosine similarity is given in Equation \eqref{eq:cosine_sim}.

Building upon cross-modal matching alignment, to enhance the model's performance in downstream visual classification tasks, we introduce a constraint cross entropy $\mathcal{L}_{CE}$ that combines image prediction results $\hat{y}$ with real labels $y$ and weights it with the matching alignment similarity constraint $\mathcal{L}_{ITA}$ for training the EMA module.
\begin{align}
    \mathcal{L}_{CE} &= -\sum_{i=1}^{N} y_i \log(\hat{y}_i) \label{loss:ce}
    \\
    \mathcal{L}_{EMA} &= {\alpha}_1 \cdot \mathcal{L}_{CE} + {\beta} \cdot \mathcal{L}_{ITA}
\end{align}
where $\alpha_1$ and $\beta$ represent constraint weights.

\subsection{Cross-Modal Diffusion Reconstruction}

In this module, we further interact the visual representations $x_v$ and semantic representations $x_s$ based on the diffusion model. Through the diffusion model, we alleviate the impact of background noise in the visual representation $x_v$ and, with the assistance of semantic representation $x_s$, generate more robust visual object representations $x_r$. In the following sections, we will first introduce the background of diffusion models and then describe the cross-modal reconstruction process based on diffusion model.
\subsubsection{Background of Diffusion Models}

The denoising diffusion probabilistic model mainly consists of two processes: a forward process $q$ with diffusion and noise addition, and a reverse process $p$ with reconstruction and denoising. In the forward process $q$, Gaussian noise is gradually added to the original training data $x$ over $T$ time steps, following a Markov process:
\begin{align}
q\left(x_t \mid x_{t-1}\right) &=\mathcal{N}\left(x_t ; \sqrt{1-\beta_t} x_{t-1}, \beta_t \mathbf{I}\right)
\\
q\left(x_t \mid x_0\right) &= \mathcal{N}\left(x_t ; \sqrt{\bar{\alpha}_t} x_0,\left(1-\bar{\alpha}_t\right) \mathbf{I}\right)
\end{align}
where $x_0 \sim q(x)$ , $\mathcal{N(.)}$ means a Gaussian distribution, $\beta_t$ determines the noise schedule. $\alpha_t = 1-\beta_t$ and $\bar{\alpha}_t = \prod_{s=0}^t \alpha_s$,  utilize the above formula to sample the noisy sample $x_t$ at any step $t$ from $x_0$.

In reverse process, the data is reconstructed by the model. The optimization objective of the model is to maximize likelihood estimation $p_{\theta}(x_0)$ of the true data distribution, where $\theta$ represents the parameters learned by a neural network.
\begin{align}
\mu_\theta\left(x_t, t\right) &= \frac{1}{\sqrt{\alpha_t}}\left(x_t-\frac{\beta_t}{\sqrt{1-\bar{\alpha}_t}} \epsilon_\theta\left(x_t, t\right)\right)
\\
p_\theta\left(x_{t-1} \mid x_t\right) &= \mathcal{N}\left(x_{t-1} ; \mu_\theta\left(x_t, t\right), \sigma_\theta\left(x_t, t\right)\right)
\end{align}

In the case of conditional guided generation, we have a data pair $(x_0, y_0) \sim (x, y)$. Similar to the above formula, we can derive:

\begin{align}
    \mu_\theta\left(x_t, t, y\right) &= \frac{1}{\sqrt{\alpha_t}}\left(x_t-\frac{\beta_t}{\sqrt{1-\bar{\alpha}_t}} \epsilon_\theta\left(x_t, t, y\right)\right)
\end{align}

\subsubsection{Cross-Modal Reconstruction}
We employ a diffusion model to reconstruct the representations extracted by the base model across modalities. Firstly, we design an multi-layer perceptron(MLP) consisting of four linear layers and activation functions for predicting $\hat{x}_s^0 = X_{\theta}(x_s^t,t,x_v)$ during the reverse process. 

In the forward process of the diffusion model, we treat the semantic representation $x_s$ as the input to the diffusion model while using the visual representation $x_v$ as a guiding condition. We smoothly interact the cross-modal representation information by gradually injecting noise. The diffusion model is to minimize the distant between $\hat{x}_s^0$ and $x_s^0$.

In this process, we construct the representation generation $\hat{x}_s^0$ constraint by calculating mean squared error (MSE) between the reconstructed semantic features $\hat{x}_s^0$ and the original input text features $x_s^0$. To enhance the performance of the generated representation on downstream tasks, similar to the Embedding Matching Align module, we introduce the cross-entropy constraint to assist in the training of the Cross-Modal Diffusion Recon module:
\begin{align}
    \mathcal{L}_{MSE} &= {\lVert X_{\theta}(x_s^t,t,x_v)-x_s^0 \rVert}_2^2
    \\
    \mathcal{L}_{CDR} &= {\alpha}_2 \cdot \mathcal{L}_{CE} + \gamma \cdot \mathcal{L}_{MSE}
\end{align}
where ${\alpha}_2$ and $\gamma$ represent constraint weights, the $\mathcal{L}_{CE}$ has been given in Equation~\eqref{loss:ce}.

Subsequently, during the reverse process, we initialize random Gaussian noise as the model input. According to Bayes theorem, $p_\theta(x_t-1|x_t)$ can be calculated according to the following definition:
\begin{align}
\hat{\beta_t} &= \frac{1-\Bar{\alpha}_{t-1}}{1-\Bar{\alpha}_t}\beta_t
\\
\mu_{\theta}\left(x_s^t,t,x_v\right) &= \frac{\sqrt{\Bar{\alpha}_{t-1}}\beta_t}{1-\Bar{\alpha}_t}X_{\theta}\left(x_s^t,t,x_v\right) + \frac{\sqrt{\alpha_t}\left(1-\Bar{\alpha}_{t-1}\right)}{1-\Bar{\alpha}_t}x_s^t
\\
p_\theta\left(x_{t-1}|x_t\right) &= \mathcal{N}\left(x_{t-1} ; \mu_{\theta}\left(x_s^t,t,x_v\right), \hat{\beta_t}I\right)
\end{align}

Similarly, guided by the visual representation $x_v$, we generate the representations $x_{CDR}$ across modalities.

\subsection{Multi-Modal Embedding Fusion}
\vspace{-0.2cm}

In the final phase, we combine the representations, $x_{EMA}$ and $x_{CDR}$, outputted by the previous two modules to realize the complementation and enhancement of information across modalities. 

Specifically, we utilize a range of techniques for the fusion of representations, encompassing direct concatenation, addition, multiplication, SUM, and Harmonic(HM)\cite{fusion_hm}. Based on the integrated representation, we conduct classification tasks, serving as MARNet's final output.
\begin{align}
    \hat{C} &= {Classifier}(x_{EMA} \oplus x_{CDR})    
\end{align}
where $\oplus$ represents the process of representation fusion, and $\hat{C}$ denotes the final prediction result.

\section{Experiment}

\subsection{Experiment Settings}

\subsubsection{Datasets}

We conducted various experiments on the task of image classification using the following two datasets:

\textbf{Vireofood-172\cite{vireo-food}:} A single-label classification dataset containing 110,241 dish images across 172 categories, including 353 textual descriptions, averages three texts per image. Following the settings in the original paper, we divided the dataset into 66,071 images for training and 33,154 images for testing.

\textbf{Ingredient-101\cite{ingredients-101}:} A single-label classification dataset comprising 93,425 dish images from the Food-101 dataset,  featuring 446 common ingredients across 101 categories, averaging 9 ingredients per dish. According to the original paper's settings, we utilized a training set consisting of 68,175 data pairs and a testing set comprising 25,250 data pairs.

\subsubsection{Performance Metrics}
Since both datasets we used are single-label datasets, we employed accuracy rate as the performance evaluation metric:
\begin{align}
    Accuracy &= \frac{TP+TN}{TP+TN+FP+FN}
\end{align}
where $TP$ is the number of true positive samples, $TN$ is the number of true negative samples, $FP$ is the number of false positive samples, and $FN$ is the number of false negative samples. For the above indicator, we calculate the average value of top-1 and top-5.

\begin{table}
\vspace{-0.3cm}
\caption{The performance results of state-of-the-art visual neural networks and alignment networks on the datasets. \textbf{Acc-1/5} refers to Accuracy Top1/5.}
\renewcommand{\arraystretch}{1.1}
\centering
\begin{tabular}{c|c|cc|cc}
\hline
\multirow{2}{*}{\textbf{Method}} & \multirow{2}{*}{\textbf{\begin{tabular}[c]{@{}c@{}}Model\end{tabular}}} & \multicolumn{2}{c|}{\textbf{Vireo-Food172}} & \multicolumn{2}{c}{\textbf{Ingredient-101}} \\ \cline{3-6} 
& & \multicolumn{1}{c|}{\textbf{Acc-1}} & \textbf{Acc-5} & \multicolumn{1}{c|}{\textbf{Acc-1}} & \textbf{Acc-5} \\ 
\hline
\multirow{10}{*}{\begin{tabular}[c]{@{}c@{}}Visual\\Classification\end{tabular}} 
 & ResNet-18       & \multicolumn{1}{c|}{77.3} & 93.2 & \multicolumn{1}{c|}{78.4} & 93.8 \\ \cline{2-6} 
 & ResNet-50       & \multicolumn{1}{c|}{81.6} & 95.0 & \multicolumn{1}{c|}{82.0} & 94.9 \\ \cline{2-6} 
 & VGG-19          & \multicolumn{1}{c|}{81.2} & 95.1 & \multicolumn{1}{c|}{81.4} & 94.3 \\ \cline{2-6} 
 & WRN             & \multicolumn{1}{c|}{82.3} & 95.5 & \multicolumn{1}{c|}{82.9} & 95.4 \\ \cline{2-6} 
 & WISeR           & \multicolumn{1}{c|}{82.8} & 96.5 & \multicolumn{1}{c|}{83.2} & 95.8 \\ \cline{2-6} 
 & RepVGG          & \multicolumn{1}{c|}{83.5} & 96.3 & \multicolumn{1}{c|}{83.6} & 96.5 \\ \cline{2-6} 
 & RepMLPNet       & \multicolumn{1}{c|}{83.3} & 96.2 & \multicolumn{1}{c|}{83.8} & 96.5 \\ \cline{2-6}
 & ViT-B/16        & \multicolumn{1}{c|}{85.4} & 97.3 & \multicolumn{1}{c|}{88.3} & 97.6 \\ \cline{2-6} 
 & ViT-B/32        & \multicolumn{1}{c|}{84.6} & 97.2 & \multicolumn{1}{c|}{87.7} & 97.6 \\ \cline{2-6}
 & Swin-T          & \multicolumn{1}{c|}{86.5} & 97.5 & \multicolumn{1}{c|}{88.6}    & 98.1    \\ \hline
% \begin{tabular}[c]{@{}c@{}}Semantic\\ Network\end{tabular} & BERT & \multicolumn{1}{c|}{98.0} & 99.9 & \multicolumn{1}{c|}{99.0} & 99.9 \\  \hline
\multirow{7}{*}{\begin{tabular}[c]{@{}c@{}}Cross-modal\\Alignment\end{tabular}} 
 & SWD &\multicolumn{1}{c|}{87.6} & 97.9 & \multicolumn{1}{c|}{88.6} & 97.7 \\ \cline{2-6} 
 & SSAN            & \multicolumn{1}{c|}{87.1} & 97.7 & \multicolumn{1}{c|}{88.5} & 97.7 \\ \cline{2-6} 
 & CDD             & \multicolumn{1}{c|}{86.0} & 97.0 & \multicolumn{1}{c|}{88.4} & 97.6 \\ \cline{2-6} 
 & SDM             & \multicolumn{1}{c|}{87.6} & 97.7 & \multicolumn{1}{c|}{88.7} & 97.7 \\ \cline{2-6} 
 & TEAM            & \multicolumn{1}{c|}{87.6} & 97.8 & \multicolumn{1}{c|}{88.7} & 97.8 \\ \cline{2-6} 
 & ITA             & \multicolumn{1}{c|}{87.8} & 97.9 & \multicolumn{1}{c|}{88.8} & 97.8 \\ \cline{2-6} 
 & \textbf{MARNet} & \multicolumn{1}{c|}{\textbf{88.1}} & \textbf{98.0} & \multicolumn{1}{c|}{\textbf{89.0}} & \textbf{97.9} \\ \hline
\end{tabular}
\label{table:image_perforamce}
\vspace{-0.3cm}
\end{table}

\subsection{Performance Analysis} 

To verify the effectiveness of our proposed MARNet in enhancing image classification performance and model robustness, we conducted experiments divided into two categories: basic visual classification and cross-modal alignment that incorporates textual information. In the visual network, we selected common structures such as Vision Transformer (ViT) and residual neural networks (ResNet). In the alignment network, we tried different alignment methods based on ViT-B/16 and BERT models, including distance measurement and similarity comparison. Table \ref{table:image_perforamce} shows the experimental results.

In experiments on visual networks, we can draw the following conclusions:
\begin{itemize}
    \item In deep convolutional neural networks, ResNet-50\cite{resnet} has more convolutional layers and a deeper network structure compared to ResNet-18\cite{resnet}, resulting in significant performance improvement. WRN\cite{wrn} and WISeR\cite{wiser} further enhance the performance of ResNet-50 by increasing the network's width and introducing feature attention mechanisms, respectively. RepVGG\cite{repvgg} and RepMLPNet\cite{repmlpnet} significantly enhance the performance of the base model VGG\cite{vgg} through structural reparameterization.

    \item Vision Transformer\cite{vit} divides the image into small patch blocks and establishes a global understanding of the image through self-attention mechanisms, achieving optimal performance. Compared to ViT-B/32 with larger patch size, ViT-B/16 can better capture the details in image information, thus achieving better performance. Swin-Transformer\cite{swin-tf} introduces hierarchical attention to enhance the quality of visual representations, further improving the model's performance.
\end{itemize}
In alignment network experiments, the following conclusions can be drawn:
\begin{itemize}
    \item In alignment methods based on distance metrics, Slice WD\cite{swd} focuses on multi-modal output spaces and performs well when semantic output is good. Simultaneous Semantic Alignment Network\cite{ssan} improves visual representation performance by attracting the centroids of representation clusters in latent space. Contrastive Domain Discrepancy\cite{cdd} uses clustering to make the clusters more compact internally while repelling each other between clusters. However, this method is susceptible to changes in initial representation quality and cluster centroid information, resulting in slightly inferior performance compared to other methods.
    \item In methods based on contrastive learning, representation pairs are divided into positive and negative samples, enhancing model attention and surpassing distance-based methods. Similarity Distribution Matching\cite{sdm}, Instance-wise Cross-modal Alignment\cite{ita}, and Token Embeddings Alignment\cite{tema} all use cosine similarity to determine sample matching degree. TEAM focuses more on positive pairs through attention, with performance heavily reliant on text quality. ITA simulates unseen negative samples within a mini-batch, increasing the distinction between positive and negative samples. Benefiting from high-quality text information in the dataset, contrastive learning-based alignment methods achieve strong and similar effects.
    \item Contrastive learning-based alignment methods aim to enhance the similarity between representation pairs in a shared latent space by distinguishing positive and negative sample pairs. Simultaneously, these methods enforce repulsion between matching image-text representation pairs and other non-matching pairs (i.e., negative samples). This alignment approach generally outperforms distance-based methods, which often overlook the negative impact caused by mismatched representation pairs within the same cluster. 
    \item Building upon contrastive learning-based matching alignment, MARNet utilizes a diffusion model guided by visual representations to generate textual representations, effectively extracting crucial textual information from images. 
    Furthermore, we strategically fuse these representations to deepen the interaction between image and text information. As a result, our innovatively proposed network achieves significantly improved performance.
\end{itemize}

\subsection{Ablation Study}

To validate the effectiveness of modules in MARNet, we conducted ablative experiments using ViT model as baseline. The results are shown in Table \ref{table:ablation}. 
\begin{itemize}
    \item Despite the potential noise interference in the image information, the baseline model demonstrates decent performance due to the detailed perception capability of ViT and the assistance of attention mechanisms. 
    \item After incorporating the EMA module and introducing high-quality textual information and alignment through contrastive matching, the model's performance improved significantly. However, the presence of residual noise and interfering factors in the visual information limits the effect of alignment. 
    \item By incorporating the CDR module, we facilitate profound interaction between visual and textual representations to derive representations founded on visual cues, thereby diminishing the impact of peripheral visual elements. Moreover, to mitigate the impact of the MLP component within the CDR module, we introduce a validation process that exclusively leverages the MLP-mapped features for assessing the efficacy of the CDR module.
    \item In the end, by integrating the EMA and CDR modules, we synergistically enhance the alignment representation and reconstruction representation, further strengthening the visual representation and model robustness.
\end{itemize}

\begin{table}
\caption{The chart presents the ablation experiment results of MARNet. \textbf{Acc-1/5} refers to the Top 1/5 Accuracy.}
\centering
\renewcommand{\arraystretch}{1.2}
\begin{tabular}{c|cc|cc}
\hline
\multirow{2}{*}{\textbf{Module}} & \multicolumn{2}{c|}{\textbf{Vireo-Food172}} & \multicolumn{2}{c}{\textbf{Ingredient-101}} \\ \cline{2-5} 
          & \multicolumn{1}{c|}{\textbf{Acc-1}} & \textbf{Acc-5} & \multicolumn{1}{c|}{\textbf{Acc-1}} & \textbf{Acc-5} \\ \hline
Baseline  & \multicolumn{1}{c|}{85.4}  & 97.3  & \multicolumn{1}{c|}{88.3}  & 97.6  \\ \hline
+EMA      & \multicolumn{1}{c|}{87.8}  & 97.9  & \multicolumn{1}{c|}{88.8}  & 97.8  \\ \hline
+MLP      & \multicolumn{1}{c|}{85.5}  & 95.4  & \multicolumn{1}{c|}{86.4}  & 94.8  \\ \hline
+CDR(MLP) & \multicolumn{1}{c|}{86.9}  & 92.5  & \multicolumn{1}{c|}{88.0}  & 90.5  \\ \hline
+Fusion   & \multicolumn{1}{c|}{\textbf{88.1}}  & \textbf{98.0}  & \multicolumn{1}{c|}{\textbf{89.0}}  & \textbf{97.9}  \\ \hline
\end{tabular}
\label{table:ablation}
\end{table}

\subsection{Case Study}

\vspace{-0.3cm}
\begin{figure}
\centering
\includegraphics[width=\textwidth]{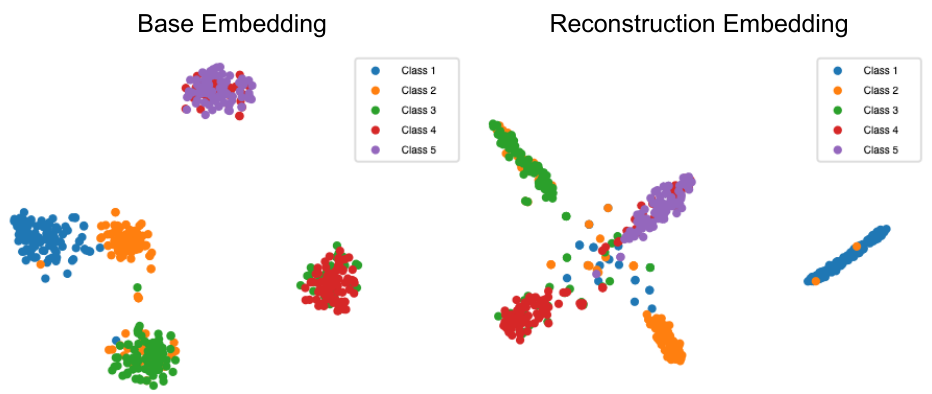}
\caption{Visualization of the basic representation $x_v$ and reconstructed representation $x_{CDR}$ using \textit{t-SNE}. As shown in the legend, the color of dots represents the category.} \label{tsne}
\end{figure}
\vspace{-0.5cm}

\subsubsection{Reconstructed feature visualization}
We conducted \textit{t-SNE} visualization on features from the ViT base model and CDR module, selecting 100 samples each from five Vireo-Food172 categories, as depicted in Figure \ref{tsne}. From the visualization results, it is evident that the reconstruction process based on the diffusion model significantly improved the distribution of the representations and effectively separated the confusing samples in the original representations. Due to the diffusion model generating based on random noise, a small number of unstable samples within the space. However, these minimal instances of noise have negligible impact on the performance of the model, as shown in Table \ref{table:ablation}.

\subsubsection{Analysis of CDR results}
\begin{figure}
\centering
\includegraphics[width=\textwidth]{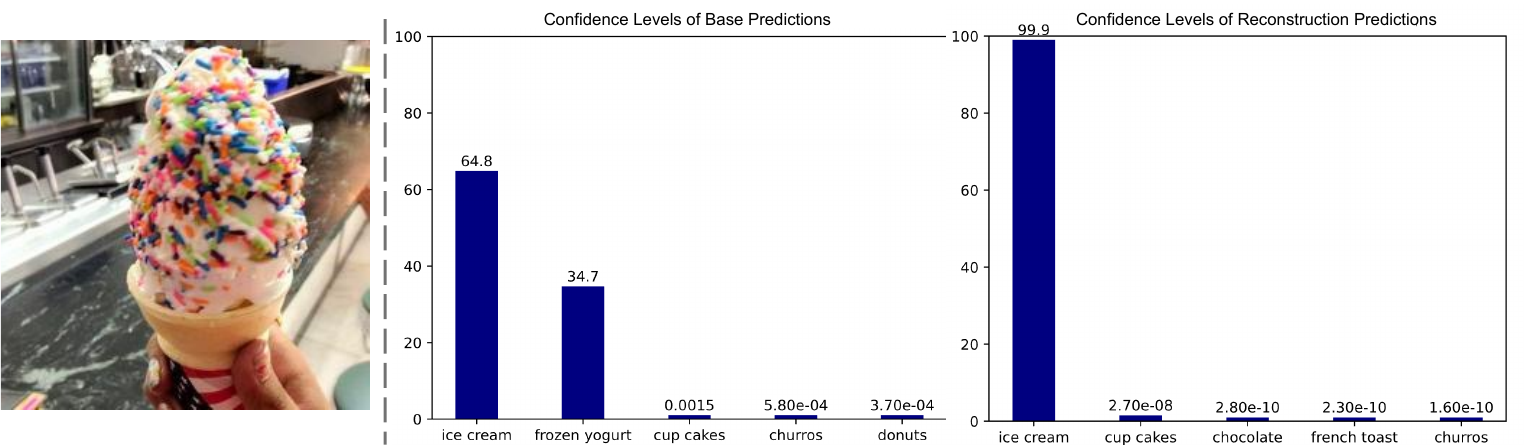}
\caption{Prediction results of the basic module and CDR module. The minimal confidence values are represented in scientific notation, e.g., 6.4e-1 indicates 0.64.} \label{recon_pred}
\vspace{-0.8cm}
\end{figure}

In ablation experiments, we can clearly observe a significant decrease in the TOP5 accuracy on the two datasets. We presented and analyzed the prediction results of base visual module and CDR module (shown in Figure \ref{recon_pred}): the confidence predicted by base model is typically distributed among top 1-3 classes. However, the diffusion model, which incorporates semantic information, tends to be completely confident in predicting a certain class, with the confidence in other classes stemming more from the randomness of sampling process. This leads to a situation where, when the prediction of the most likely class is incorrect, base model can maintain a relatively high TOP5 accuracy, while the diffusion model struggles to improve.

\section{Conclusion}
In this article, we tackle the issue of heterogeneity in multi-modal data by introducing the Multi-Modal Alignment and Reconstruction Network (MARNet). This network addresses the disparities in distance and distribution within the feature space through a dual approach: embedding matching alignment(EMA) modules and cross-modal diffusion reconstruction(CDR) modules. Our experimental findings validate that MARNet significantly improves the quality of visual information and optimizes the distribution of representations.

Moving forward, our efforts will concentrate on reducing noise interference during reconstruction phase of the diffusion model, with the overarching goal of preserving the integrity of original information to the greatest extent possible.

\section*{Acknowledgments}
This work is supported in part by the Oversea Innovation Team Project of the  "20 Regulations for New Universities" funding program of Jinan (Grant no. 2021GXRC073)

\bibliographystyle{splncs04}
\begin{scriptsize}
\bibliography{reference}  
\end{scriptsize}

\end{document}